\documentclass[sigconf]{acmart}
\AtBeginDocument{%
  }

\copyrightyear{2026}
\acmYear{2026}
\setcopyright{cc}
\setcctype{by}
\acmConference[MM '26]{Proceedings of the 34th ACM International Conference on Multimedia}{November 10--14, 2026}{Rio de Janeiro, Brazil}
\acmBooktitle{Proceedings of the 34th ACM International Conference on Multimedia (MM '26), November 10--14, 2026, Rio de Janeiro, Brazil}
\acmDOI{10.1145/3767308.3836229}
\acmISBN{979-8-4007-2213-4/2026/11}

\usepackage{booktabs}
\usepackage{multirow}
\usepackage{subcaption}

\begin{document}

\title{Efficient Chain-of-Modality Reasoning via Progressive Compression for Spoken Language Models}

\author{Pengchao Feng}
\authornote{Work done during an internship at Token Foundry, Alibaba Group. }
\email{the.bird@sjtu.edu.cn}
\orcid{0009-0000-1469-9543}
\affiliation{%
  \institution{Shanghai Jiao Tong University
  Shanghai Innovation Institute}
  \city{Shanghai}
  \country{China}
}

\author{Chao-Hong Tan}
\email{chtan@mail.ustc.edu.cn}
\orcid{0009-0004-8589-0408}
\affiliation{%
  \department{Token Foundry}
  \institution{Alibaba Group}
  \city{Shanghai}
  \country{China}
}

\author{Qian Chen}
\email{lukechan1231@gmail.com}
\orcid{0000-0001-6939-7438}
\affiliation{%
  \department{Token Foundry}
  \institution{Alibaba Group}
  \city{Hangzhou}
  \country{China}
}

\author{Wen Wang}
\email{wwang.969803@gmail.com}
\orcid{0000-0002-0356-1968}
\affiliation{%
  \department{Token Foundry}
  \institution{Alibaba Group}
  \city{Sunnyvale}
  \state{CA}
  \country{USA}
}

\author{Xiangang Li}
\email{lixiangang.lxg@alibaba-inc.com}
\orcid{0009-0003-7164-935X}
\affiliation{%
  \department{Token Foundry}
  \institution{Alibaba Group}
  \city{Hangzhou}
  \country{China}
}

\author{Xie Chen}
\email{chenxie95@sjtu.edu.cn}
\orcid{0000-0001-7423-617X}
\correspondingauthor
\affiliation{%
  \institution{Shanghai Jiao Tong University
  Shanghai Innovation Institute}
  \city{Shanghai}
  \country{China}
}

\renewcommand{\shortauthors}{Pengchao Feng et al.}

\begin{abstract}
Spoken language models (SLMs) enable natural human-computer interaction, but their reasoning ability still lags behind that of text-based large language models, especially on spoken mathematical question answering tasks. One important reason is that SLMs reason over purely verbalized mathematical expressions, which are harder to interpret than symbolic text. However, directly transferring text-based reasoning to SLMs is nontrivial due to architectural constraints and the additional computational requirements. To address this challenge, we propose Efficient Chain-of-Modality Reasoning (ECoM Reasoning), the first framework to introduce compressed reasoning into SLMs. By compressing the textual component so that it jointly serves as speech guidance and reasoning representation, ECoM Reasoning improves reasoning accuracy while using a smaller token budget than the standard Chain-of-Modality (CoM) architecture, which generates intermediate text before speech. To train this capability, we further propose Progressive Compression, a curriculum-based strategy that gradually trains the model from full-form reasoning to compressed reasoning. Experiments on spoken mathematical question answering benchmarks show that ECoM Reasoning improves accuracy by 21\% over standard CoM without explicit reasoning, and by 3\% over CoM with full reasoning traces while using only 40\% of the text tokens, demonstrating that it enhances SLM reasoning while remaining inference-efficient.

\end{abstract}



\maketitle

\section{Introduction}
\footnote{Project page: \url{https://github.com/QwenAudio/FunResearch}} Spoken language models (SLMs) are emerging as a promising foundation for natural and expressive human-computer interaction. By directly modeling speech as both input and output, they enable richer interaction patterns than text-only systems, including more natural turn-taking~\cite{hurst2024gpt}, prosody-aware responses~\cite{wang2024blsp, xue2024chat}, and direct spoken communication~\cite{defossez2024moshi, yu2024salmonn}. Current SLMs mainly adopt three types of speech generation architectures: (1) Parallel generation, which generates text and speech outputs in parallel~\cite{xu2025qwen3, ding2025kimi, fang2025llama, chen2025slam, xie2024mini, zhang2025mimo, zhao2025moss}; (2) Interleaved generation, which interleaves text and speech tokens within a single output stream~\cite{zeng2024glm, li2025baichuan}; and (3) Chain-of-Modality (CoM), which generates user text, assistant text, and assistant speech in sequence~\cite{zhang2023speechgpt, zhang2024speechgpt, arora2025chain, arora2025chain2}. Recent advances in end-to-end speech modeling have substantially improved the quality of speech understanding and generation, suggesting that SLMs may become an important interface for future intelligent agents.

However, strong speech interaction does not necessarily translate into strong reasoning~\cite{chiang2025stitch}. This gap is especially evident on tasks requiring precise and structured reasoning, where SLMs remain weaker than text-based LLMs, particularly for multi-step spoken mathematical question answering~\cite{lin2025voice, wei2025towards}. A key challenge is that SLMs must reason over purely verbalized mathematical expressions (e.g., "x plus five y equals zero"), which are much harder to interpret than symbolic text~\cite{hyeon2025mathspeech, xie2025mini}. 

\begin{figure*}[t]
    \centering
    \includegraphics[width=0.96\linewidth]{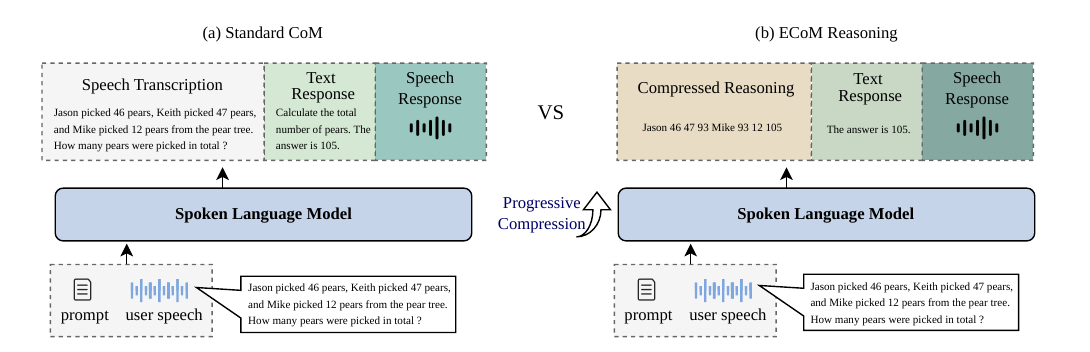}
    \caption{Overview of (a) the \textsc{standard CoM} framework and (b) the \textsc{ECoM Reasoning} framework, both built upon the Chain-of-Modality (CoM) architecture. By compressing the textual component, \textsc{ECoM Reasoning} enables the intermediate text to simultaneously guide speech generation and carry the core reasoning process, while maintaining low inference cost.}
    \label{fig:framework}
\end{figure*}

A natural way to mitigate this difficulty is to introduce explicit intermediate textual reasoning. Yet directly inserting textual reasoning before the model response is not straightforward, as it is constrained by the underlying architecture. In interleaved or parallel speech-text generation architectures, adding textual reasoning would require extensive padding in the speech channel, substantially increasing computational cost and delaying the first useful speech token; it may also cause speech to be generated earlier than the corresponding text, leading to misaligned or unstable outputs. In contrast, Chain-of-Modality (CoM)~\cite{zhang2023speechgpt,arora2025chain}, in which the model first generates text and then produces speech conditioned on it (as shown in Fig.~\ref{fig:framework}(a)), provides a more suitable architecture for incorporating reasoning text. However, CoM also exposes a key inefficiency: stronger reasoning typically requires longer textual reasoning traces, which directly increase inference cost. This trade-off is especially problematic for SLMs because text already functions as an intermediate modality for spoken response generation. As a result, simply improving reasoning by generating more text is not an ideal solution. A central question, then, is whether the textual component in SLMs can simultaneously support speech generation and reasoning with fewer tokens. We argue that this is possible if the model learns to use text not as a full-form reasoning transcript, but as a compact reasoning carrier that preserves only the essential information needed for problem solving and response generation.


Based on this idea, we propose \textbf{Efficient Chain-of-Modality Reasoning} (\textsc{ECoM Reasoning}), to the best of our knowledge, the first end-to-end speech interaction framework to introduce compressed reasoning into SLMs (as shown in Fig.~\ref{fig:framework}(b)). Instead of generating full-form CoM-style textual reasoning outputs, \textsc{ECoM Reasoning} compresses the textual component so that it serves two functions simultaneously: guiding speech generation while encoding the key reasoning information needed to solve the task. In this way, the text modality becomes both a speech-guiding representation and an efficient reasoning interface. By increasing the functional density of the textual component, \textsc{ECoM Reasoning} improves the reasoning ability of SLMs with a smaller overall inference budget.

Another key challenge is how to train such a speech interaction framework to perform compressed reasoning effectively. Directly enforcing the model to learn short reasoning traces can hurt output quality, as it is difficult for the model to simultaneously learn spoken understanding, speech generation, and reasoning over compressed text. Inspired by curriculum learning~\cite{bengio2009curriculum, hammoud2025train}, we argue that the model should learn to reason fully before learning to reason compactly. Based on this intuition, we propose \textbf{Progressive Compression}, a curriculum-based training strategy in which the model first learns basic spoken understanding and generation, then full reasoning in spoken dialogue, and finally compressed reasoning. In this way, the model gradually acquires the ability to preserve essential reasoning information in end-to-end speech interaction.

We evaluate \textsc{ECoM Reasoning} on spoken mathematical question answering benchmarks (AddSub~\cite{hosseini2014learning}, SingleEq~\cite{koncel2015parsing}, MultiArith~\cite{roy2015solving}, and SVAMP~\cite{patel2021nlp}), which require both speech understanding and precise reasoning. \textsc{ECoM Reasoning} improves accuracy by 21\% over standard CoM without explicit reasoning, and by 3\% over CoM with full reasoning traces while using \textbf{only 40\%} of the text tokens. Across benchmarks, our method achieves the best accuracy-per-token trade-off, demonstrating that compressed reasoning can improve the reasoning efficiency of SLMs with lower inference cost. These results suggest that stronger spoken reasoning does not necessarily require larger inference budgets, but can instead be achieved through a more efficient and functionally dense use of text.

In short, our main contributions are as follows:
\begin{itemize}
    \item We propose \textbf{ECoM Reasoning}, to the best of our knowledge, the first framework to introduce compressed reasoning into SLMs, enabling the textual modality to jointly support speech generation and reasoning with a smaller inference budget.
    \item We propose \textbf{Progressive Compression}, a curriculum-based training strategy that gradually guides the model from full-form reasoning to compressed reasoning.
    \item We demonstrate on spoken mathematical  question answering benchmarks that \textsc{ECoM Reasoning} outperforms both \textsc{standard CoM} and \textsc{CoM Reasoning} while using fewer text tokens, achieving \textbf{the best accuracy-per-token trade-off}.
\end{itemize}

\section{Related Work}
\subsection{Spoken Language Model Reasoning}
Related work on SLM reasoning has developed along two directions: \textit{(i)} improving the reasoning capability of SLMs on speech and audio tasks, and \textit{(ii)} improving the engineering realization of reasoning under real-time spoken interaction. 

The first direction focuses on mitigating the gap between the waveform signal and language semantics of reasoning~\cite{wang2025multimodal}. For understanding and translation, CoT-ST~\cite{du2024cot} and Xie et al.~\cite{xie2025leveraging} pipeline acoustic processing with downstream semantic tasks, while Audio-CoT ~\cite{ma2025audio} and Audio-Reasoner ~\cite{xie2025audio} introduce structured, multi-stage reasoning for general audio understanding and long-horizon tasks. In generative settings, SpatialSonic~\cite{sun2024both} and SpeechGPT-Gen~\cite{zhang2024speechgpt} employ step-wise attribute extraction and information chaining to progressively align acoustic features with high-level semantic planning. TARS~\cite{wang2026closing} proposes systematic alignment strategies to harmonize acoustic perception with linguistic reasoning in SLMs.

The second direction focuses on making reasoning compatible with low-latency spoken interaction. Recent studies explore simultaneous or incremental processing, where models reason during streaming input rather than after the full utterance is observed~\cite{shih2025can}. Representative examples include chunked or simultaneous reasoning frameworks such as STITCH~\citep{chiang2025stitch} and SHANKS~\citep{chiang2025shanks}. There is also work on reasoning for cascading dialogue systems, including input-time speculation~\citep{li2025predgen}, listen-think-speak frameworks~\citep{zou2026lts}, dual-brain reasoning~\citep{wu2025mind}, and dual-track streaming response architectures~\citep{liu2026discourse}.

While prior work has successfully adapted textual reasoning paradigms to SLMs, it has largely overlooked the efficiency of the reasoning process itself. To address this limitation, we pioneer the integration of \textbf{reasoning compression} into SLMs, opening a new direction for building more efficient reasoning and interaction frameworks for spoken language models.

\subsection{Efficient Reasoning}

Recent work on efficient reasoning seeks to retain the benefits of Chain-of-Thought (CoT) while reducing the latency and computational cost of long reasoning traces~\cite{sui2025stop}. Broadly, existing methods fall into two categories: training-based approaches that teach models to reason more compactly, and training-free approaches that control reasoning more efficiently at inference time.

Training-based methods improve reasoning efficiency by shortening or compressing intermediate reasoning during learning. Prior work studies distilling explicit CoT into implicit reasoning~\cite{deng2023implicit}, progressively internalizing CoT supervision~\cite{deng2024explicit, shen2025er}, and more generally framing implicit reasoning as an alternative to verbose rationale generation~\cite{li2025implicit}. More recent approaches use curriculum learning~\cite{bengio2009curriculum} to move from full reasoning to compressed reasoning under progressively tighter token budgets~\cite{hammoud2025train}, or compress reasoning more adaptively through token-level skipping, dynamic stopping, and self-distillation~\cite{chen2025aware, sang2026policy, li2026making}. Other work compresses reasoning processes in latent space by going beyond next-token prediction~\cite{tan2025think}.

Training-free methods instead reduce unnecessary reasoning at test time without modifying model parameters. These include using short draft-style reasoning in place of full CoT~\cite{xu2025chain}, shifting reasoning from discrete tokens to continuous concept representations~\cite{zhang2025soft, zhuang2025text}, and allocating token budgets adaptively based on problem complexity~\cite{lin2025plan}. Overall, these studies suggest that reasoning efficiency can be improved either by training models to produce more compact rationales or by adaptively constraining reasoning during inference.

However, most existing studies focus on text-based LLMs and pay little attention to the efficiency of reasoning in spoken language models. Our work fills this gap by bringing efficient reasoning ideas, particularly curriculum-based compression, into the SLM setting.

\section{Method}
\subsection{ECoM Reasoning Framework}
\textbf{Efficient Chain-of-Modality Reasoning} (\textsc{ECoM Reasoning}) is built upon the \textbf{Chain-of-Modality} (CoM) architecture (Fig.~\ref{fig:framework}). By compressing the textual component, it enables text to jointly serve as speech guidance and a carrier of core reasoning, thereby reducing inference cost.

\subsubsection{\textbf{Chain of Modality.}}

For clarity of exposition, we describe our method in the setting of a single-turn multimodal dialogue. Let \(S\) denote the system prompt, \(X^s\) the user speech input, and \(Y^s\) the assistant speech response. Our goal is to model the posterior distribution of the assistant speech response conditioned on the available user inputs and the system prompt, i.e., \(P(Y^s \mid X^s, S)\). To explicitly incorporate structured reasoning, we introduce the intermediate textual variables \(X^t\) and \(Y^t\), where \(X^t\) denotes the transcript of the user speech \(X^s\), and \(Y^t\) represents the assistant's textual response before speech generation. Then we can expand the posterior as
\begin{equation}
\begin{aligned}
P(Y^s \mid X^s, S)
&=\sum_{X^t, Y^t} P(Y^s, Y^t, X^t \mid X^s, S) \\
&=\sum_{X^t, Y^t} \bigg[ P(Y^s \mid Y^t, X^t, X^s, S)\, 
   \\&  \cdot P(Y^t \mid X^t,  X^s, S)\, \cdot P(X^t \mid X^s, S)\bigg ].
\end{aligned}
\end{equation}

Since exact marginalization is intractable in practice, we adopt a Viterbi-style approximation:
\begin{equation}
P(Y^s \mid X^s, S)
\approx
P(Y^s \mid \hat{Y}^t, \hat{X}^t, X^s, S),
\end{equation}
where
\begin{equation}
\begin{aligned}
& \hat{X}^t = \arg\max_{X^t} P(X^t \mid X^s, S), \\
& \hat{Y}^t = \arg\max_{Y^t} P(Y^t \mid \hat{X}^t, X^s, S).
\end{aligned}
\end{equation}
Under this approximation, the model first predicts a textual representation of the user input, then generates the assistant text conditioned on it, and finally produces the assistant speech. In other words, given input speech \(X^s\), the interaction process can be modeled as a sequential generation chain \(X^s \rightarrow X^t \rightarrow Y^t \rightarrow Y^s\) (as shown in Fig.~\ref{fig:framework}(a)), which is the \textsc{standard CoM}~\cite{zhang2023speechgpt}.
\begin{equation}
\textsc{standard CoM}: \quad
[S]\,[X^s] \;\rightarrow\; [X^t]\,[Y^t]\,[Y^s].
\end{equation}
where each part is separated by a special token.
This paradigm has been shown to provide both semantic coherence and high-quality synthesized speech~\cite{zhang2024speechgpt, arora2025chain, arora2025chain2}, and it naturally extends to multi-turn settings by conditioning on the dialogue history and applying the same response generation process at each turn.

To further enable explicit reasoning, we introduce an additional intermediate textual variable \(R^t\), which represents the model's reasoning process prior to response generation. The posterior can then be approximated as
\begin{equation}
P(Y^s \mid X^s, S)
\approx
P(Y^s \mid \hat{Y}^t, \hat{R}^t, \hat{X}^t, X^s, S),
\end{equation}
where
\begin{equation}
\begin{aligned}
\hat{X}^t &= \arg\max_{X^t} P(X^t \mid X^s, S), \\
\hat{R}^t &= \arg\max_{R^t} P(R^t \mid \hat{X}^t, X^s, S), \\
\hat{Y}^t &= \arg\max_{Y^t} P(Y^t \mid \hat{R}^t, \hat{X}^t, X^s, S).
\end{aligned}
\end{equation}

Under this formulation, the model first transcribes the user speech into text, then generates an intermediate reasoning trace, next produces the final textual response conditioned on that reasoning, and finally synthesizes the assistant speech, yielding the following generation chain:
\begin{equation}
\textsc{CoM Reasoning}: \quad
[S]\,[X^s] \;\rightarrow\; [X^t]\,[R^t]\,[Y^t]\,[Y^s].
\end{equation}

\subsubsection{\textbf{Compressed Data Construction.}}

To enable reasoning without increasing the overall text-token inference cost, we compress the intermediate user text \(X^t\) at the sentence level and compress \(R^t\) at the token level according to a predefined compression ratio. The compressed reasoning representation is denoted by \(\tilde{R}^t\).

\textbf{Sentence-level compression of \(X^t\).}
For the intermediate user text \(X^t\), we adopt an aggressive sentence-level compression. Specifically, the entire user text with its surrounding special tokens is removed from the training sequence. This design is motivated by our empirical finding that progressive training enables the model to retain strong speech understanding ability regardless of whether reasoning is included, even when \(X^t\) is fully removed from the output sequence. 

\textbf{Token-level compression of \(R^t\).}
For the reasoning text \(R^t\), we apply token-level compression. Specifically, we first compute an importance score for each reasoning token, and then retain only a certain proportion of the most important tokens while removing the rest, producing a shorter compressed reasoning sequence \(\tilde{R}^t\). We adopt this relatively conservative strategy because we observe that fully removing \(R^t\) causes a much larger degradation in performance than full reasoning, while selective token preservation better balances reasoning effectiveness and inference efficiency.

To estimate token importance, we adopt \textbf{LLMLingua-2}~\cite{pan2024llmlingua} as the token-importance scorer, following TokenSkip~\cite{xia2025tokenskip}. We choose this method because it does not require any modification to the underlying LLM architecture, and can achieve a higher reduction rate of the reasoning steps~\cite{sui2025stop}. Specifically, given an input token sequence \(x_{\le n} = (x_1, \ldots, x_n)\), we feed the full sequence into LLMLingua-2 (the parameters are denoted as \(\theta_{\mathrm{MB}}\)) and the importance of the \(i\)-th token \(x_i\) is calculated as
\begin{equation}
I_2(x_i) = P(x_i \mid x_{\le n}; \theta_{\mathrm{MB}}),
\end{equation}
Here, \(I_2(x_i)\) reflects the estimated importance of token \(x_i\) conditioned on the entire input sequence, since LLMLingua-2 is trained as a token-level classifier with supervision derived from token-importance annotations. We then use these scores to retain high-importance tokens and remove less informative ones during compression.

To verify that this token-level compression strategy is suitable for our data, we manually sampled and inspected compressed examples from our dataset. We provide representative cases on the project page, which show that the method preserves the key reasoning content well after compression.

In summary, by applying compression at both the sentence level and the token level, we construct the training data for the compressed reasoning model. An example of this data construction process is shown in Table~\ref{tab:compression_data}. Within the training setup described in Section~\ref{sec:training_setting}, this compression procedure is applied to the mathematical reasoning data to create compressed reasoning supervision, whereas the general dialogue data is used in its original form for spoken dialogue training.

\begin{table}[h]
\centering
\caption{Example of compressed data construction. \texttt{<u\_t>}, \texttt{<r\_t>}, and \texttt{<a\_t>} are special tokens indicating the user text, reasoning text, and assistant text segments, respectively, and \texttt{<eot>} marks the end of a segment.}
\label{tab:compression_data}
\begin{tabular}{l p{5.9cm}}
\toprule
\textbf{Compression} & \textbf{Example Sequence} \\
\midrule
Full sequence 
& ``\texttt{<u\_t>} If I have two apples and buy one more, how many apples do I have? \texttt{<eot>} \texttt{<r\_t>} The user asks a simple counting question. Start with two apples and add one more. \texttt{<eot>} \texttt{<a\_t>} You have three apples. \texttt{<eot>}'' \\

Compress \(X^t\) 
& ``\texttt{<r\_t>} The user asks a simple counting question. Start with two apples and add one more. \texttt{<eot>} \texttt{<a\_t>} You have three apples. \texttt{<eot>}'' \\

Compress \(R^t\) 
& ``\texttt{<r\_t>} two apples one more \texttt{<eot>} \texttt{<a\_t>} You have three apples. \texttt{<eot>}'' \\
\bottomrule
\end{tabular}
\end{table}

\subsubsection{\textbf{Efficient Chain-of-Modality Reasoning.}}
After compressed data construction, we provide a compact approximation to the same posterior objective while requiring fewer generated tokens:
\begin{equation}
P(Y^s \mid X^s, S)
\approx
P(Y^s \mid \hat{Y}^t, \hat{\tilde{R}}^t, X^s, S),
\end{equation}
where
\begin{equation}
\begin{aligned}
\hat{\tilde{R}}^t &= \arg\max_{\tilde{R}^t} P(\tilde{R}^t \mid X^s, S), \\
\hat{Y}^t &= \arg\max_{Y^t} P(Y^t \mid \hat{\tilde{R}}^t, X^s, S).
\end{aligned}
\end{equation}
Compared with explicitly modeling \(X^t\) and \(R^t\), this formulation preserves the essential reasoning signal while improving token efficiency.

Based on the probabilistic formulation above, we provide the sequence realization:
\begin{equation}
\textsc{ECoM Reasoning}: \quad
[S]\,[X^s] \;\rightarrow\; [\tilde{R}^t]\,[Y^t]\,[Y^s],
\end{equation}

\subsection{Progressive Training Pipeline}

\begin{figure*}[t]
    \centering
    \includegraphics[width=0.95\linewidth]{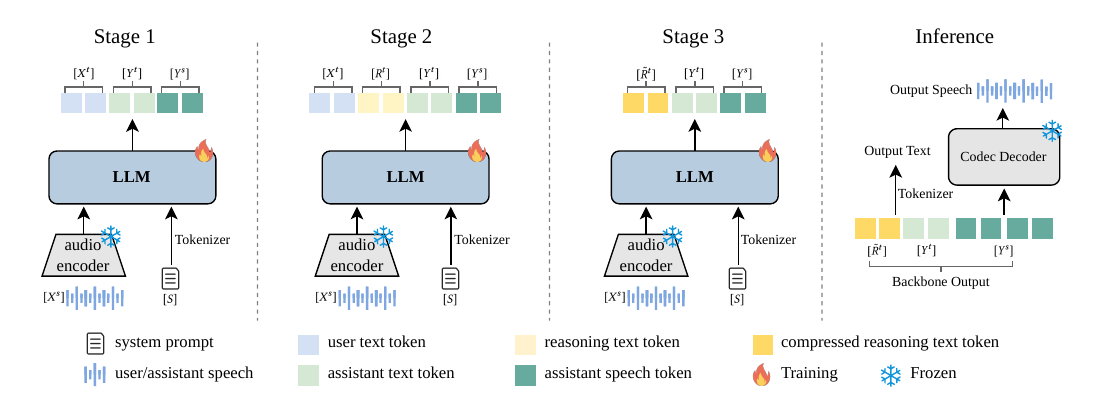}
    \caption{Overview of the proposed progressive training pipeline for \textsc{ECoM Reasoning}. The model is trained in three stages, transitioning from standard speech interaction to full-form reasoning supervision and finally to compressed reasoning.}
    \label{fig:pipeline}
\end{figure*}

Directly training \textsc{ECoM Reasoning} is difficult because the model must simultaneously learn speech interaction and compressed textual reasoning. Such joint optimization is highly challenging for the backbone model, especially when the compressed text is expected to both support speech generation and carry the essential reasoning process. To address this issue, we adopt a curriculum-style progressive training strategy that moves from easier objectives to harder ones. As shown in Fig.~\ref{fig:pipeline}, the proposed pipeline consists of three stages: \textsc{Standard CoM}, \textsc{CoM Reasoning}, and \textsc{ECoM Reasoning}.

\paragraph{\textbf{Stage 1: Standard CoM}}
We first train the model in the standard \textsc{CoM} setting on speech dialogue and spoken question answering datasets that do not contain explicit reasoning supervision, allowing the model to acquire the basic ability to understand spoken input and generate spoken output.

\paragraph{\textbf{Stage 2: CoM Reasoning}}
We then introduce a \textsc{CoM Reasoning} stage with explicit supervision over the full-form intermediate reasoning chain.
At this stage, training data includes full reasoning traces, which help the model learn the complete reasoning process before compression is introduced.

\paragraph{\textbf{Stage 3: ECoM Reasoning}}
Finally, we transition to \textsc{ECoM Reasoning}, where the full intermediate text is replaced by a compact reasoning representation. This enables the model to preserve reasoning capability while reducing the cost of text-token generation at inference time.

In summary, the progressive pipeline first builds fundamental speech interaction ability, then injects explicit reasoning supervision, and finally distills this capability into a more efficient compressed-text reasoning format.

\section{Experiments}

\subsection{Experimental Setup}
\paragraph{\textbf{Training Setting.}}
\label{sec:training_setting}
We use two categories of training data: general dialogue data for developing the model's spoken dialogue ability, and mathematical reasoning data for enhancing its reasoning capability on math-intensive tasks. 

For general dialogue training, we use \textit{MagpiePro}~\cite{xu2024magpie} and \textit{InfGen}~\cite{li2025infinity}, with speech synthesized by CosyVoice3~\cite{du2025cosyvoice}, following DrVoice~\cite{tan2025drvoice} and Fun-Audio-Chat~\cite{team2025fun}. We keep only single-turn dialogue examples and discard samples with combined speech and text token length exceeding 4K. After filtering, the resulting dialogue corpus contains 12.7K hours of speech.

For mathematical reasoning training, we use GSM8K~\cite{cobbe2021training}, NuminaMath~\cite{li2024numinamath}, and MathQA~\cite{amini2019mathqa}. To ensure that each sample contains an explicit reasoning process, we keep only samples with \texttt{question}, \texttt{solution}, and \texttt{answer} fields, thereby excluding proof-style problems and other samples without step-by-step solutions. We then synthesize speech using GPT-4o-mini-TTS~\cite{hurst2024gpt}. The resulting math reasoning corpus contains 206,216 samples in total, corresponding to 1.5K hours of synthesized speech.

Our training framework is built on top of SLAM-LLM~\cite{ma2026slam}, as it provides a unified and modular foundation for speech-language modeling, naturally supporting the integration of speech encoding, intermediate text/reasoning generation, and speech decoding in our framework. The SLM used in our experiments consists of a Whisper encoder~\cite{radford2023robust}, a CosyVoice 3 decoder~\cite{du2025cosyvoice}, and Qwen2.5-1.5B~\cite{qwen2.5, qwen2}. In addition, following SLAM-Omni~\cite{chen2025slam}, we adopt group modeling for speech-token decoding, which reduces the effective length of speech token sequences and improves decoding efficiency.

\paragraph{\textbf{Evaluation Setting.}}
We evaluate our method on both reasoning and non-reasoning benchmarks. Our primary evaluation focuses on reasoning tasks. To this end, we consider four math reasoning datasets: AddSub~\cite{hosseini2014learning}, SingleEq~\cite{koncel2015parsing}, MultiArith~\cite{roy2015solving}, and SVAMP~\cite{patel2021nlp}, which are widely used to assess the mathematical reasoning ability of LLMs. The questions and answers in these datasets are of moderate difficulty and can be naturally expressed in speech, making them suitable for evaluating SLMs of moderate scale. For spoken response evaluation, we use the same complete evaluation set as STITCH~\cite{chiang2025stitch}, including the synthesized speech items.

In addition, we evaluate general spoken dialogue ability using a subset of non-reasoning test sets selected from  UltraEval-Audio~\cite{shi2026ultraeval}. These test sets include Llama Questions~\cite{nachmani2023spoken}, a knowledge-oriented generated-speech dataset. 

All experiments are conducted under the URO-Bench and UltraEval-Audio evaluation frameworks.

\subsection{Main Results}
\begin{table*}[t]
\caption{Results on math question benchmarks. We compare a cascade system, the standard \textsc{CoM} architecture, \textsc{CoM Reasoning} with full and budget-constrained reasoning, and our proposed \textsc{ECoM Reasoning}. We report accuracy (\%)~$\uparrow$, the average number of generated text tokens (\#Tok)~$\downarrow$, and token efficiency (Acc/\#Tok)~$\uparrow$. The best result in each column is shown in \textbf{bold}.}
\label{tab:reasoning}
\centering
\resizebox{\ifdim\width>0.99\textwidth 0.99\textwidth\else\width\fi}{!}{%
\begin{tabular}{lccccccccccccccc}
\toprule
\multirow{2}{*}{Model}
& \multicolumn{3}{c}{AddSub}
& \multicolumn{3}{c}{MultiArith}
& \multicolumn{3}{c}{SingleEq}
& \multicolumn{3}{c}{SVAMP}
& \multicolumn{3}{c}{\textbf{Average}} \\
\cmidrule(lr){2-4}\cmidrule(lr){5-7}\cmidrule(lr){8-10}\cmidrule(lr){11-13}\cmidrule(lr){14-16}
& Acc~$\uparrow$ & \#Tok~$\downarrow$ & Acc/\#Tok~$\uparrow$
& Acc~$\uparrow$ & \#Tok~$\downarrow$ & Acc/\#Tok~$\uparrow$
& Acc~$\uparrow$ & \#Tok~$\downarrow$ & Acc/\#Tok~$\uparrow$
& Acc~$\uparrow$ & \#Tok~$\downarrow$ & Acc/\#Tok~$\uparrow$
& Acc~$\uparrow$ & \#Tok~$\downarrow$ & Acc/\#Tok~$\uparrow$ \\
\midrule

Cascade 
&53.21	&72.48	&0.73	
&58.04	&111.82	&0.52	
&61.16	&78.56	&0.78	
&\textbf{48.82}	&109.58	&0.45	
&55.31	&93.11	&0.62 \\

\textsc{standard CoM}
& 46.78 & 166.04 & 0.28
& 43.67 & 188.18 & 0.23
& 34.55 & 159.30 & 0.22
& 33.33 & 203.01 & 0.16
& 39.58 & 179.13 & 0.22 \\

\textsc{CoM Reasoning}
& 48.62 & 100.85 & 0.48
& 63.40 & 95.41 & 0.66
& 72.78 & 86.92 & 0.84
& 46.15 & 103.07 & 0.45
& 57.74 & 96.56 & 0.61 \\

\quad w/ budget = 20
& 21.10 & 29.86 & 0.71
& 0.57 & 33.00 & 0.02
& 26.91 & 30.15 & 0.89
& 10.81 & 29.36 & 0.37
& 14.85 & 30.59 & 0.50 \\

\quad w/ budget = 30
& 50.15 & 41.10 & 1.22
& 5.17 & 39.48 & 0.13
& 40.06 & 40.44 & 0.99
& 23.29 & 39.61 & 0.59
& 29.67 & 40.16 & 0.73 \\

\midrule
\textsc{ECoM Reasoning}
& \textbf{52.59} & \textbf{31.92} & \textbf{1.65}
& \textbf{71.83} & \textbf{31.05} & \textbf{2.31}
& \textbf{75.84} & \textbf{26.02} & \textbf{2.91}
& 42.36 & \textbf{31.86} & \textbf{1.33}
& \textbf{60.66} & \textbf{30.21} & \textbf{2.05} \\
\bottomrule
\end{tabular}
}
\end{table*}

\begin{table*}[t]
\caption{Ablation on the reasoning-text compression ratio. We compress the reasoning text \(R^t\) to \(100\%\), \(80\%\), \(60\%\), \(40\%\), \(20\%\), and \(0\%\) of its original length. We report accuracy (\%)~$\uparrow$, the average number of generated text tokens (\#Tok), and token efficiency (Acc/\#Tok)~$\uparrow$. The best result in each column is shown in \textbf{bold}.}
\label{tab:cr}
\centering
\resizebox{\ifdim\width>0.99\textwidth 0.99\textwidth\else\width\fi}{!}{%
\begin{tabular}{cccccccccccccccc}
\toprule
\textsc{ECoM Reasoning}
& \multicolumn{3}{c}{AddSub}
& \multicolumn{3}{c}{MultiArith}
& \multicolumn{3}{c}{SingleEq}
& \multicolumn{3}{c}{SVAMP}
& \multicolumn{3}{c}{\textbf{Average}} \\
\cmidrule(lr){2-4}\cmidrule(lr){5-7}\cmidrule(lr){8-10}\cmidrule(lr){11-13}\cmidrule(lr){14-16}
Compress To
& Acc~$\uparrow$ & \#Tok~$\downarrow$ & Acc/\#Tok~$\uparrow$
& Acc~$\uparrow$ & \#Tok~$\downarrow$ & Acc/\#Tok~$\uparrow$
& Acc~$\uparrow$ & \#Tok~$\downarrow$ & Acc/\#Tok~$\uparrow$
& Acc~$\uparrow$ & \#Tok~$\downarrow$ & Acc/\#Tok~$\uparrow$
& Acc~$\uparrow$ & \#Tok~$\downarrow$ & Acc/\#Tok~$\uparrow$ \\
\midrule

\(100\%\)
& 47.09 & 72.58 & 0.65
& 64.36 & 57.68 & 1.12
& 66.05 & 57.65 & 1.15
& 41.24 & 65.18 & 0.63
& 54.69 & 63.27 & 0.89 \\

\(80\%\)
& 42.81 & 44.89 & 0.95
& 66.09 & 45.67 & 1.45
& 68.80 & 41.50 & 1.66
& 42.36 & 50.27 & 0.84
& 55.02 & 45.58 & 1.23 \\

\(60\%\)
& 52.29 & 37.23 & 1.40
& \textbf{72.22} & 36.34 & 1.99
& 72.47 & 30.31 & 2.39
& \textbf{43.03} & 37.52 & 1.15
& 60.00 & 35.35 & 1.73 \\

\(40\%\)
& 52.59 & 31.92 & 1.65
& 71.83 & 31.05 & 2.31
& \textbf{75.84} & 26.02 & 2.91
& 42.36 & 31.86 & 1.33
& \textbf{60.66} & 30.21 & 2.05 \\

\(20\%\)
& \textbf{54.43} & 26.07 & 2.09
& 54.59 & 22.37 & \textbf{2.44}
& 68.19 & 21.12 & 3.23
& 35.78 & 23.44 & 1.53
& 53.25 & 23.25 & 2.32 \\

\(0\%\)
& 44.03 & \textbf{10.21} & \textbf{4.31}
& 14.36 & \textbf{9.00} & 1.60
& 58.71 & \textbf{9.39} & \textbf{6.25}
& 23.18 & \textbf{9.03} & \textbf{2.57}
& 35.07 & \textbf{9.41} & \textbf{3.68} \\
\bottomrule
\end{tabular}
}
\end{table*}

Table~\ref{tab:reasoning} reports the main results on math reasoning benchmarks. We compare four frameworks: a cascade system built from the same backbone components (ASR: Whisper-small, LLM: Qwen2.5-1.5B, and TTS: CosyVoice3), the standard \textsc{CoM} architecture, \textsc{CoM Reasoning}, and our proposed \textsc{ECoM Reasoning}. For \textsc{CoM Reasoning}, we further include two inference-time budget-constrained variants, in which reasoning generation is truncated once the token budget is reached, forcing the model to proceed directly to text and speech response generation. For \textsc{ECoM Reasoning}, we report the variant with reasoning compressed to 40\% of the original length.

We evaluate them with three metrics: (1) \textbf{Accuracy (Acc)}, which measures whether the model correctly answers the math problem; following URO-Bench~\cite{yan2025uro}, we transcribe the generated speech with Whisper-large-v3 and then use GPT-mini for evaluation; 
(2) \textbf{Average Generated Text Tokens (\#Tok)}, which measures the average number of text tokens generated by the model; we focus on text tokens instead of speech tokens because our study investigates whether the budget saved by compressing text can be used for reasoning; before reporting this metric, we remove extreme outliers using an interquartile range (IQR) based criterion with the threshold coefficient set to 10; and 
(3) \textbf{Token Efficiency (Acc/\#Tok)}, defined as accuracy divided by the average number of generated text tokens, which reflects how effectively the model uses text tokens~\cite{chen2025aware}.

Several observations can be made. First, standard \textsc{CoM} performs the worst, with an average accuracy of only 39.58, indicating that explicit reasoning is essential for spoken mathematical problem solving. Second, \textsc{ECoM Reasoning} achieves higher accuracy than \textsc{CoM Reasoning} while using substantially fewer text tokens. On average, it improves accuracy from 57.74 to 60.66, while reducing the number of generated text tokens from 96.56 to 30.21. It also delivers the best token efficiency, reaching 2.05, compared with 0.61 for \textsc{CoM Reasoning}, 0.62 for the cascade system, and 0.22 for standard \textsc{CoM}. Finally, simply imposing a tight token budget on \textsc{CoM Reasoning} at inference time severely harms reasoning performance, especially on multi-step problems such as MultiArith. This result further shows that our method improves the reasoning capability of SLMs under the same inference cost.

We further repeat the same experiments under streaming inference mode to measure the latency of the first speech token (FST). All inference experiments are conducted on a single A800 GPU. \textsc{CoM Reasoning} yields an average first-token latency of 4.90\,s, whereas \textsc{ECoM Reasoning} reduces it to 1.60\,s, demonstrating the practical feasibility of our framework for real-time spoken interaction.

\subsection{Ablation and Analysis}
\subsubsection{\textbf{Effect of Compression Ratio}}

We further study the effect of the reasoning-text compression ratio in compressed data construction. Specifically, we compress the reasoning text \(R^t\) to \(100\%\), \(80\%\), \(60\%\), \(40\%\), \(20\%\), and \(0\%\) of its original length, and train/evaluate all models under the same configuration. Table~\ref{tab:cr} reports the results. We find that a compression ratio of \(40\%\) achieves the best overall performance, yielding the highest response accuracy and high token efficiency. Interestingly, although the \(0\%\) setting attains the highest token efficiency, its response accuracy drops substantially, indicating that it is not the optimal choice. This also suggests that the Acc/\#Tok metric has limitations as a measure of efficiency, since a higher value does not necessarily correspond to better overall reasoning quality.

This finding indicates that the original reasoning text contains substantial redundancy. Moderate compression removes non-essential tokens while preserving the core reasoning process, whereas insufficient compression retains unnecessary token cost and excessive compression risks discarding useful reasoning information.

Table~\ref{tab:compression_case} presents a representative example. The full-form reasoning (\(100\%\)) is verbose but correct, while compressed variants at \(80\%\), \(60\%\) and \(40\%\) preserve only the key arithmetic structure and still lead to the correct answer. However, the \(20\%\) setting produces an incorrect intermediate reasoning path and thus yields a wrong answer, suggesting that excessive compression can lead to reasoning errors. Overall, these results support the use of moderate reasoning compression in \textsc{ECoM Reasoning}.

\begin{table*}[t]
\centering
\caption{Example of reasoning-text compression under different compression ratios. Question: ``A bookstore had 41 books in the bargain bin. If they sold 33 books but then put 2 more in the bin, how many books would be in the bin?''}
\label{tab:compression_case}
\begin{tabular}{c p{12.8cm} c}
\toprule
\textbf{Ratio} & \textbf{Predict Text} & \textbf{Answer} \\
\midrule

\textsc{CoM Reasoning}
& A bookstore had 41 books in the back room. If they sold 33 books but then put 2 more in the bin, how many books would be in the bin? The bookstore had \(41 - 33 = 8\) books left in the back room. After putting 2 more in the bin, there would be \(8 + 2 = 10\) books in the bin. The answer is 10.
& 10 \\
\midrule

\textsc{ECoM Reasoning}
\\
100\%
& The number of books in the bin is \(41 - 33 = 8\) books. After adding 2 books, the number of books in the bin is \(8 + 2 = 10\) books. The answer is 10.
& 10 \\
\midrule

80\%
& bargain bin \(41 - 33 = 8\) books. Adding 2 books, bin \(8 + 2 = 10\) books. The answer is 10.
& 10 \\
\midrule

60\%
& bookstore sold 33 books \(41 - 33 = 8\) books put 2 books \(8 + 2 = 10\) books. The answer is 10.
& 10 \\
\midrule

40\%
& bookstore sold 33, \(41 - 33 = 8\), put 2, \(8 + 2 = 10\). The answer is 10.
& 10 \\
\midrule

20\%
& \(41\;\;33\;\;74\;\;2\;\;76\). The answer is 76.
& 76 \\
\midrule

0\%
& The answer is 10.
& 10 \\
\bottomrule
\end{tabular}
\end{table*}

\textbf{Why can moderate compression improve answer accuracy?}
Simply having a longer thought process does not guarantee improved results~\cite{balachandran2025inference}. One possible explanation is that unconstrained reasoning often contains substantial redundancy, and excessively long generated sequences are more susceptible to distraction and error accumulation~\cite{sang2026policy}. In contrast, concise reasoning may offer a cleaner and more effective signal for prediction.

\subsubsection{\textbf{Impact of Training Strategy}}




 

\begin{table*}[t]
\caption{Ablation on progressive training strategies for \textsc{ECoM Reasoning}, comparing one-stage, two-stage, three-stage (reversed), four-stage, and our default three-stage curriculum. We report accuracy (\%)~$\uparrow$, the average number of generated text tokens (\#Tok)~$\downarrow$, and token efficiency (Acc/\#Tok)~$\uparrow$. The best result in each column is shown in \textbf{bold}.}
\label{tab:pt}
\centering
\resizebox{\ifdim\width>0.99\textwidth 0.99\textwidth\else\width\fi}{!}{%
\begin{tabular}{lccccccccccccccc}
\toprule
\multirow{2}{*}{Strategy}
& \multicolumn{3}{c}{AddSub}
& \multicolumn{3}{c}{MultiArith}
& \multicolumn{3}{c}{SingleEq}
& \multicolumn{3}{c}{SVAMP}
& \multicolumn{3}{c}{\textbf{Average}} \\
\cmidrule(lr){2-4}\cmidrule(lr){5-7}\cmidrule(lr){8-10}\cmidrule(lr){11-13}\cmidrule(lr){14-16}
& Acc~$\uparrow$ & \#Tok~$\downarrow$ & Acc/\#Tok~$\uparrow$
& Acc~$\uparrow$ & \#Tok~$\downarrow$ & Acc/\#Tok~$\uparrow$
& Acc~$\uparrow$ & \#Tok~$\downarrow$ & Acc/\#Tok~$\uparrow$
& Acc~$\uparrow$ & \#Tok~$\downarrow$ & Acc/\#Tok~$\uparrow$
& Acc~$\uparrow$ & \#Tok~$\downarrow$ & Acc/\#Tok~$\uparrow$ \\
\midrule
One-stage
& 36.69 & 76.14 & 0.48
& 11.30 & 35.64 & 0.32
& 37.61 & \textbf{9.50} & \textbf{3.96}
& 23.85 & \textbf{9.19} & \textbf{2.60}
& 27.36 & 32.62 & 1.84 \\

Two-stage
& 35.47 & \textbf{25.59} & 1.39
& 50.38 & \textbf{28.76} & 1.75
& 61.46 & 26.87 & 2.29
& 37.23 & 29.50 & 1.26
& 46.14 & \textbf{27.68} & 1.67 \\

Three-stage (reversed)
& 53.51 & 28.16 & 1.90
& 53.44 & 28.32 & 1.89
& 63.60 & 27.23 & 2.34
& 41.80 & 30.87 & 1.35
& 53.09 & 28.65 & 1.87 \\

Four-stage
& \textbf{57.49} & 28.17 & \textbf{2.04}
& 61.49 & 29.18 & 2.11
& 69.41 & 26.87 & 2.58
& 38.23 & 29.86 & 1.28
& 56.66 & 28.52 & 2.00 \\

\textbf{Three-stage (ours)}
& 52.59 & 31.92 & 1.65
& \textbf{71.83} & 31.05 & \textbf{2.31}
& \textbf{75.84} & 26.02 & 2.91
& \textbf{42.36} & 31.86 & 1.33
& \textbf{60.66} & 30.21 & \textbf{2.05} \\
\bottomrule
\end{tabular}
}
\end{table*}

We further ablate the progressive training strategy by comparing several alternative training schedules for \textsc{ECoM Reasoning}. Specifically, we evaluate: 
(1) a one-stage setting that directly trains \textsc{ECoM Reasoning} from scratch; 
(2) a two-stage setting that trains \textsc{CoM} first and then directly transitions to \textsc{ECoM Reasoning}; 
(3) a three-stage variant with reversed compression order, where compressed reasoning is introduced before removing the user text; and 
(4) a four-stage setting that separately compresses the user text and the reasoning text in two successive stages. 
To ensure a fair comparison, we keep the remaining training hyperparameters unchanged across all settings, including the initial learning rate, total training steps, and batch size, despite the different numbers of stages.

The results are reported in Table~\ref{tab:pt}. Our default three-stage strategy consistently achieves the best response accuracy, suggesting that it provides the most effective curriculum for transferring the benefits of textual guidance and explicit reasoning into the compressed format. By comparison, fewer training stages reduce token usage during generation but lead to inferior reasoning performance, indicating insufficient adaptation to the compressed representation. In contrast, the four-stage strategy brings no further improvement and may introduce a greater risk of overfitting.

\subsubsection{\textbf{Performance on Non-reasoning SpeechQA}}
For non-reasoning tasks, we trained the \textsc{ECoM} interaction architecture without explicit reasoning, following the same progressive compression principle, whose format is
\begin{equation}
\textsc{ECoM}: \quad
[S]\,[X^s] \;\rightarrow\; [X^t]\,[Y^t]\,[Y^s] \;\Rightarrow\; [Y^t]\,[Y^s].
\end{equation}
And we compare three settings: \textsc{CoM}, directly trained \textsc{ECoM}, and progressively trained \textsc{ECoM}. As shown in Table~\ref{tab:llamaq}, \textsc{CoM} attains slightly better response accuracy, while the progressively trained \textsc{ECoM} model achieves much higher token efficiency. This suggests that compressed intermediate text can retain most of the useful guidance for knowledge-based speechQA while significantly reducing the cost of text generation.

\begin{table}[t]
\caption{Results on the No-reasoning SpeechQA. We report accuracy (\%)~$\uparrow$, the average number of generated text tokens (\#Tok)~$\downarrow$, and token efficiency (Acc/\#Tok)~$\uparrow$. }
\label{tab:llamaq}
\centering
\small
\begin{tabular}{lccc}
\toprule
\multirow{2}{*}{Model w/o reasoning} & \multicolumn{3}{c}{LlamaQ} \\
\cmidrule(lr){2-4}
& Acc~$\uparrow$ & \#Tok~$\downarrow$ & Acc/\#Tok~$\uparrow$ \\
\midrule
standard CoM    & \textbf{56.00} & 153.21 & 0.37 \\
ECoM (direct) & 49.66 & 84.95 & 0.58 \\
ECoM (progressive)   & 52.66 & \textbf{80.68} & \textbf{0.65} \\
\bottomrule
\end{tabular}
\end{table}

\textbf{Why does progressive training help?}

We perform a case study to analyze the mechanism behind progressive training. Specifically, we compare two \textsc{ECoM} models: one trained directly from scratch and one trained progressively, starting from an initial \textsc{CoM} stage. For the same input instance, we examine token-level generation and measure the similarity between the last-layer hidden states of the \textsc{ECoM} model and those of a \textsc{CoM} model at the corresponding token positions.

The corresponding results are shown in Fig.~\ref{fig:layer_similarity_llamaqa}. \textsc{ECoM}-progressive exhibits consistently higher similarity than \textsc{ECoM}-direct across all 29 layers (average gain of $0.058$), with non-overlapping 95\% confidence intervals. This indicates that progressive training enables the model to better capture the full textual generation chain before transitioning to the compressed format. Consequently, although the final generated text is compressed, the associated hidden representations still preserve information from the original full-form text sequence.

\subsubsection{\textbf{Impact of Compressing Assistant Text}}
\begin{figure}[t]
    \centering
    \begin{subfigure}[t]{0.48\linewidth}
        \centering
        \includegraphics[width=\linewidth]{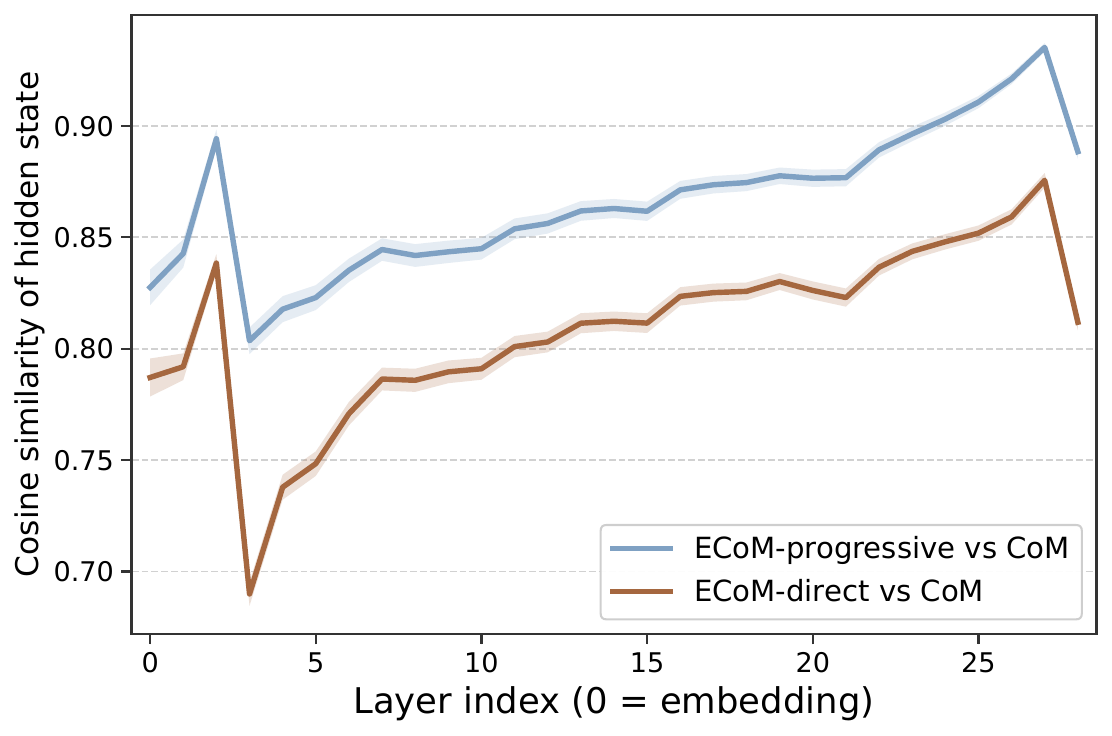}
        \caption{Layer-wise hidden-state similarity to \textsc{CoM} on Llama Questions. \textsc{ECoM}-progressive shows consistently higher similarity than \textsc{ECoM}-direct across all 29 layers, with non-overlapping 95\% confidence intervals.}
        \label{fig:layer_similarity_llamaqa}
    \end{subfigure}
    \hfill
    \begin{subfigure}[t]{0.48\linewidth}
        \centering
        \includegraphics[width=\linewidth]{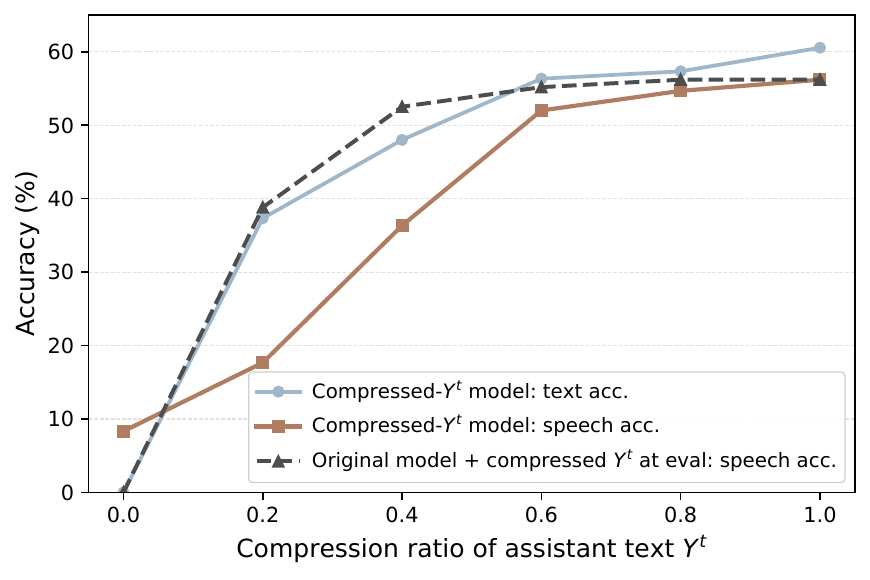}
        \caption{Accuracy under different retained ratios when compressing the assistant text \(Y^t\) on Llama Questions. We compare models trained with compressed \(Y^t\) and an original model evaluated with inference-only compression.}
        \label{fig:assistant_compression}
    \end{subfigure}
    \caption{Case studies on progressive training and assistant-text compression.}
    \label{fig:case_study_llamaqa}
\end{figure}

\textbf{Why is the assistant text \(Y^t\) not compressed?}
We observe that, unlike \(X^t\) and \(R^t\), the assistant text \(Y^t\) is highly sensitive to compression. To study this effect, we evaluate two scenarios across different retained ratios of \(Y^t\): 
(1) a model trained and evaluated with compressed assistant text, and 
(2) an original model for which \(Y^t\) is compressed only at inference time. The results are presented in Fig.~\ref{fig:assistant_compression}.

Notably, if compressed \(Y^t\) remained sufficient for recovering the full speech response, then the inference-only compression setting should still allow the model to produce competitive speech outputs. In practice, however, both settings suffer clear degradation in speech accuracy as the retained ratio decreases, with especially severe deterioration at aggressive compression levels. This shows that \(Y^t\) is not merely a redundant textual byproduct, but a crucial intermediate representation that is tightly coupled with downstream speech generation. Therefore, while compressing user text and reasoning text is effective, finding a lossless compression strategy for assistant text remains an open problem.

\subsubsection{\textbf{Impact of Different Architectures}}
\textbf{Why do we adopt the CoM-based architecture?}
Using the general dialogue training data, the same backbone and training configurations, we pre-train three end-to-end variants: \textsc{Parallel}, \textsc{Interleave}, and \textsc{standard CoM} without reasoning, and compare them with the cascade system built from the same backbone components.

Results on the LLaMA Questions benchmark show that \textsc{standard CoM} achieves 56.00\% accuracy, substantially outperforming \textsc{Interleave} (45.67\%) and \textsc{Parallel} (38.33\%). Moreover, the pre-trained \textsc{CoM} model approaches the cascade system, which reaches 60.33\% accuracy.

In addition, our method requires inserting relatively long reasoning spans before the final assistant response, which is poorly suited to the generation patterns of \textsc{Interleave} and \textsc{Parallel}. By contrast, \textsc{CoM} naturally accommodates such structured insertion. Therefore, we use \textsc{CoM} as the backbone architecture throughout this work.

\section{Limitations}
\textsc{ECoM Reasoning} improves the reasoning efficiency of spoken language models without increasing inference cost, but several limitations remain.
First, similar to the Chain-of-Modality (\textsc{CoM}) architecture, it requires generating text before speech, leading to higher time-to-first-speech-token latency than parallel or interleaved frameworks; however, this remains acceptable for accuracy-critical tasks such as spoken mathematical QA or tutoring. 
Second, we only evaluate on a 1.5B model, leaving scalability to larger backbones unverified, though our results already suggest strong potential. 
Third, our evaluation is limited to English, and cross-lingual or multilingual settings remain for future work.

\section{Conclusion}
We introduce \textsc{ECoM Reasoning}, to our knowledge, the first framework that incorporates compressed textual reasoning into spoken language models. The key idea is to rethink the role of intermediate text in SLMs: rather than serving only as a scaffold for speech generation, it can be compressed into an efficient representation that also carries the core reasoning process. Built on the CoM architecture, this design enables better use of the text-token budget without increasing inference cost. To make such a representation learnable, we further propose a progressive compression strategy that gradually bridges full-form reasoning and compact reasoning. We believe this opens a new and promising avenue for future research on more intelligent spoken language models.

\begin{acks}
This work was supported by National Natural Science Foundation of China  (No. U23B2018), the Science and Technology Innovation (STI) 2030-Major Project (2022ZD0208700), Yangtze River Delta Science and Technology Innovation Community Joint Research Project (2024CSJGG1100), and Alibaba Research Intern Program.
\end{acks}


\bibliographystyle{ACM-Reference-Format}
\bibliography{refs}

\clearpage
\appendix
\section{Training Configuration}

As a supplement to Section~4.1 (Experimental Setup), Table~\ref{tab:training_config} summarizes the detailed training configuration. All experiments were conducted on 4 NVIDIA A800 GPUs using distributed data parallel training with \texttt{torchrun}. We used FP16 mixed precision and performed full-parameter training.

We initialized the language model with Qwen2.5-1.5B and the speech encoder with Whisper-Small. The hidden dimensions were 1536 and 768, respectively. Input features were 80-bin mel-spectrograms. For audio tokenization, we used CosyVoice 3 codec tokens. Group decoding was enabled with a linear adapter, where one token was projected into three tokens (i.e., group size = 3).

Following common practice in recent SLM training (e.g., SLAM-Omni), our setup is as follows: (i) the LLM backbone jointly generates text and speech tokens under a unified next-token cross-entropy objective; (ii) the speech encoder is frozen, while the encoder adapter is trained in all three stages; and (iii) the speech-token projection layer is trained throughout, while the codec decoder remains frozen. Importantly, all three stages use both speech input and speech output, so the training remains speech-coupled throughout. 

We trained with a batch size of 2 per GPU for 2 epochs (each epoch approximately 63{,}900 steps), using a learning rate of $1\times10^{-5}$. We used a warmup-and-cosine decay schedule with 3{,}000 warmup steps and 300{,}000 total training steps. Validation was performed every 6{,}390 steps with a 1\% split, so that 10 checkpoints corresponded to one epoch. The speech encoder was frozen during training, while the LLM remained trainable.

\begin{table}[h]
\centering
\small
\caption{Training configuration used in our experiments.}
\label{tab:training_config}
\begin{tabular}{ll}
\toprule
\textbf{Setting} & \textbf{Value} \\
\midrule
Hardware & 4 $\times$ NVIDIA A800 GPUs \\
Distributed Training & DDP (\texttt{torchrun}) \\
Precision & FP16 \\
PEFT & Disabled \\
\midrule
LLM Backbone & Qwen2.5-1.5B \\
LLM Hidden Size & 1536 \\
Speech Encoder & Whisper-Small \\
Encoder Hidden Size & 768 \\
Input Feature & 80-bin mel-spectrogram \\
Audio Code Type & CosyVoice 3 \\
Group Decoding & Enabled \\
Group Size & 3 \\
Group Decode Adapter & Linear \\
\midrule
Batch Size / GPU & 2 \\
Effective Batch Size & 8 \\
Epochs & 2 \\
Learning Rate & $1\times10^{-5}$ \\
Warmup Steps & 3000 \\
Max Steps & 300000 \\
Validation Interval & 6390 steps \\
Validation Split & 0.01 \\
Encoder Freeze & Yes \\
\bottomrule
\end{tabular}
\end{table}

\section{Prompt Configuration}

We use the same system prompt for all subsets to maintain a simple and consistent training setup and to facilitate fair comparisons across experiments.

\noindent \textbf{System Prompt:} \\
Conduct a spoken conversation with the user.

\section{Evaluation Configuration}
As a supplement to Section~4.1, Table~\ref{tab:eval_config} summarizes the evaluation configuration. All results were obtained using a unified test-time template, and the same decoding settings were applied across experiments to ensure fair comparisons.

For decoding, we used deterministic generation with a text repetition penalty of 1.0 and an audio repetition penalty of 1.2, which helps reduce silence. The maximum generation length was set to 9000 tokens. The audio prompt was fixed to the default tone for all evaluations.

\begin{table}[h]
\centering
\small
\caption{Unified evaluation configuration used in all experiments.}
\label{tab:eval_config}
\begin{tabular}{ll}
\toprule
\textbf{Setting} & \textbf{Value} \\
\midrule
Test Split & \texttt{test} \\
Dataset Version & \texttt{speech-to-speech} \\
Random Seed & 888 \\
\midrule
Text Repetition Penalty & 1.0 \\
Audio Repetition Penalty & 1.2 \\
Max New Tokens & 9000 \\
Sampling & Disabled \\
Output Text Only & False \\
Speech Sample Rate & 24000 \\
Online Inference & Disabled \\
Audio Prompt & \texttt{default\_tone} \\
\bottomrule
\end{tabular}
\end{table}

\section{Case Studies}

We provide audio-annotated case studies from the main experiments and ablation studies on the project page for further inspection.

\begin{table*}[htp]
\caption{Comparison of efficient-reasoning methods. \#Eff: Acc/\#Tok~}
\label{tab:reasoning_new}
\centering
\resizebox{\ifdim\width>0.9\textwidth 0.9\textwidth\else\width\fi}{!}{%
\begin{tabular}{lccccccccccccccc}
\toprule
\multirow{2}{*}{Method}
& \multicolumn{3}{c}{AddSub}
& \multicolumn{3}{c}{MultiArith}
& \multicolumn{3}{c}{SingleEq}
& \multicolumn{3}{c}{SVAMP}
& \multicolumn{3}{c}{\textbf{Average}} \\
\cmidrule(lr){2-4}\cmidrule(lr){5-7}\cmidrule(lr){8-10}\cmidrule(lr){11-13}\cmidrule(lr){14-16}
&
Acc~$\uparrow$ & \#Tok~$\downarrow$ & $\uparrow$
& Acc~$\uparrow$ & \#Tok~$\downarrow$ & \#Eff~$\uparrow$
& Acc~$\uparrow$ & \#Tok~$\downarrow$ & \#Eff~$\uparrow$
& Acc~$\uparrow$ & \#Tok~$\downarrow$ & \#Eff~$\uparrow$
& Acc~$\uparrow$ & \#Tok~$\downarrow$ & \#Eff~$\uparrow$ \\
\midrule
\textbf{ECoM Reasoning(Ours)}
& 52.59 & 31.92 & 1.65
& 71.83 & 31.05 & 2.31
& 75.84 & 26.02 & 2.91
& 42.36 & 31.86 & 1.33
& \textbf{60.66} & 30.21 & 2.05\\

LLMLingua-1
& 48.01 & 61.16 & 0.78
& 73.94 & 55.25 & 1.34
& 76.14 & 48.62 & 1.57
& 40.80 & 60.96 & 0.67
& 59.72 & 56.50 & 1.09 \\

LongLLMLingua
& 46.78 & 61.00 & 0.77
& 70.49 & 53.44 & 1.32
& 69.72 & 47.11 & 1.48
& 42.02 & 59.68 & 0.70
& 57.25 & 55.31 & 1.07 \\

Chain-of-Draft
& 41.28 & 56.02 & 0.74
& 66.47 & 54.82 & 1.21
& 53.82 & 45.57 & 1.18
& 34.11 & 57.77 & 0.59
& 48.92 & 53.55 & 0.93 \\

Cold-Stop
& 36.08 & 33.41 & 1.08
& 14.55 & 34.00 & 0.43
& 38.53 & 30.76 & 1.25
& 28.98 & 30.47 & 0.95
& 29.54 & 32.16 & 0.93 \\

Stepwise Internalisation
& 39.44 & 9.52 & 4.14
& 10.34 & 9.00 & 1.15
& 49.23 & 9.43 & 5.22
& 27.31 & 9.00 & 3.03
& 31.58 & \textbf{9.24} & \textbf{3.39} \\

Heima
& 35.77 & 11.83 & 3.02
& 8.62 & 11.00 & 0.78
& 50.76 & 11.33 & 4.48
& 29.20 & 11.00 & 2.65
& 31.09 & 11.29 & 2.74 \\

\bottomrule
\end{tabular}
}
\end{table*}

\begin{table*}[htp]
\caption{Speech synthesis quality on the four math-QA benchmarks.}
\label{tab:tts_quality}
\centering
\resizebox{\ifdim\width>0.9\textwidth 0.9\textwidth\else\width\fi}{!}{%
\begin{tabular}{lcccccccccc}
\toprule
\multirow{2}{*}{Method}
& \multicolumn{2}{c}{AddSub}
& \multicolumn{2}{c}{MultiArith}
& \multicolumn{2}{c}{SingleEq}
& \multicolumn{2}{c}{SVAMP}
& \multicolumn{2}{c}{Average} \\
\cmidrule(lr){2-3}\cmidrule(lr){4-5}\cmidrule(lr){6-7}\cmidrule(lr){8-9}\cmidrule(lr){10-11}
& WER$\%\downarrow$ & UTMOS$\uparrow$
& WER$\%\downarrow$ & UTMOS$\uparrow$
& WER$\%\downarrow$ & UTMOS$\uparrow$
& WER$\%\downarrow$ & UTMOS$\uparrow$
& WER$\%\downarrow$ & UTMOS$\uparrow$ \\
\midrule
LFM2-Audio
& \textbf{11.04} & 4.11 & 12.68 & 4.13 & 13.09 & 4.13 & \phantom{0}9.75 & 4.15 & 11.64 & 4.13 \\
Cascade
& 17.94 & 4.14 & 12.47 & 4.21 & \phantom{0}8.93 & 4.24 & 11.04 & 4.20 & 12.60 & 4.20 \\
\midrule

CoM
& 19.46 & \textbf{4.37} & 10.52 & \textbf{4.37} & 14.80 & \textbf{4.38} & 13.65 & \textbf{4.32} & 14.61 & \textbf{4.36} \\
CoM Reasoning
& 13.79 & 4.25 & 11.37 & 4.18 & 10.14 & 4.22 & 10.03 & 4.22 & 11.33 & 4.22 \\
ECoM Reasoning
& 14.14 & 4.17 & \textbf{10.40} & 4.13 & \textbf{\phantom{0}7.43} & 4.12 & \textbf{\phantom{0}9.53} & 4.15 & \textbf{10.38} & 4.14 \\
\bottomrule
\end{tabular}
}
\end{table*}

\section{Other Efficient-reasoning Methods} 

Broader comparisons would help provide a more comprehensive assessment of our method. In addition to the \textbf{budget-constrained} baseline reported in the main paper, we further re-implemented several representative efficient-reasoning methods using the same backbone and training setting. These include \textbf{LLMLingua-1}~\cite{jiang-etal-2023-llmlingua} and \textbf{LongLLMLingua}~\cite{jiang-etal-2024-longllmlingua}, which use token-level importance scoring as a drop-in replacement for our LLMLingua-2 scorer on the reasoning text $R^t$; \textbf{Chain-of-Draft}~\cite{xu2025chain} and \textbf{Cold-Stop} (proposed in~\cite{zhang2025soft}), which use prompts or model-entropy signals, respectively, to guide early exiting from the reasoning trace to the answer span without additional training; and \textbf{Stepwise Internalisation}~\cite{deng2024explicit} and \textbf{Heima}~\cite{shen2025er}, which distill the reasoning process into latent states or special tokens through curriculum learning. 

Since none of these methods has been previously applied to SLMs, all implementations were adapted by ourselves and may therefore differ from the original works in certain hyperparameter choices and implementation details. As shown in Table~\ref{tab:reasoning_new}, our method remains competitive among these extended baselines. In particular, implicit-CoT methods~\cite{deng2024explicit, shen2025er} compress the reasoning process more aggressively and thus achieve higher token efficiency, but at the cost of lower accuracy. The advantage of our method stems from more accurate selection of key reasoning tokens than~\cite{jiang-etal-2023-llmlingua, jiang-etal-2024-longllmlingua}, stronger training--inference consistency than~\cite{xu2025chain, zhang2025soft}, and better preservation of intermediate reasoning signals than~\cite{deng2024explicit, shen2025er}.

\section{Speech-quality evaluation} 
Although the main focus of this work is the correctness of spoken answers under a constrained token budget, speech naturalness is also important. We evaluate speech quality using two metrics: \textbf{WER} (between the assistant text and speech) and \textbf{UTMOS}. For context, we also report results for an open-source 1.5B SLM, LFM2-Audio-1.5B~\cite{liquidai2025lfm2}, and for a cascade system. As shown in Table~\ref{tab:tts_quality}, ECoM Reasoning shows only a small drop in UTMOS (4.14), which is also close to the reported human speech score on LibriSpeech-PC (4.09)~\cite{chen-etal-2025-f5}, suggesting limited impact on speech generation quality.

\section{AI Usage Statement}

AI-based tools were used only for language polishing and proofreading. They were not used for experimental design, data processing, model development, result analysis, or scientific content generation. All technical contributions and conclusions are the responsibility of the authors.

\end{document}